# Effective Integration of Symbolic and Connectionist Approaches through a Hybrid Representation

**Marcio Moreno, Daniel Civitarese, Rafael Brandao, Renato Cerqueira**
IBM Research, Brazil
{mmoreno, sallesd, rmello, rcerq}@br.ibm.com

**Abstract**

In this paper, we present our position for a neural-symbolic integration strategy, arguing in favor of a hybrid representation to promote an effective integration. Such description differs from others fundamentally, since its entities aim at representing AI models in general, allowing to describe both non-symbolic and symbolic knowledge, the integration between them and their corresponding processors. Moreover, the entities also support representing workflows, leveraging traceability to keep track of every change applied to models and their related entities (e.g., data or concepts) throughout the lifecycle of the models.

## 1 Introduction

The dualism between the approaches of connectionist and symbolic in artificial intelligence has regularly been addressed in the literature. Hilario [1995], Sun and Alexandre [1997], and Garcez et al. [2002] discuss how integrating these two approaches (neural-symbolic integration) may suppress their shortcomings.

Hilario [1995] splits the various approaches to neural-symbolic integration into two types of strategy: unified and hybrid. The former are strategies that aim at combining neural networks with symbolic capabilities that are part of the neural structures and processes alone. The latter are strategies that rely on a synergistic combination of neural and symbolic models. According to Hilario, this hybrid combination can be either translational or functional. Translational hybrids are systems where the neural networks act as processors, like the ones in the unified strategies, but there are translations between the symbolic structures and the neural networks. Functional hybrids comprise the symbolic structures, neural networks, and their corresponding processors.

Hilario has also identified four integration modes for the functional hybrids, which vary according to how symbolic and neural subsystems cooperate. Figure 1 illustrates these integration modes. In the chain processing mode, one of the subsystems (neural or symbolic) is the central processor, while the other handles the pre-processing or post-processing tasks. In the sub-processing mode, one of the two subsystems is embedded and subordinate to the other, which acts as the primary problem solver. In the meta-processing mode, one subsystem is the first level problem solver, and the other is a mid-level function compared to the first. In the coprocessing mode, the subsystems are equal partners in the problem-solving process.

Garcez et al. [2002] also discuss a way of positioning these approaches. A drawback of these works (Hilario and Garcez) is the lack of dynamicity to the solutions, being capable of switching between modes or even using only one of the subsystems. An execution engine (see Figure 1), capable of understanding and executing integration descriptions, according to a specific representation, should act as a middleware to promote this dynamicity, manipulating the symbolic and neural subsystems when appropriate.

In this paper, we present our position for a neural-symbolic integration strategy, arguing in favor of a hybrid representation to promote effective integration of neural and symbolic subsystems, capable of manipulating these subsystems in different modes in a broader sense. That is, providing abstractions to handle symbolic structures, neural networks, their corresponding processors while defining the relationships between these entities. This hybrid representation was introduced by Moreno et al. [2017] but focusing on how the hypermedia fundamentals could increase the expressivity of knowledge representations.

The remainder of this paper is organized as the following. Section 2 discusses the main entities of the proposed representation. We model a scenario using the proposed representation in Section 3. Finally, Section 4 concludes with some considerations and reflections.

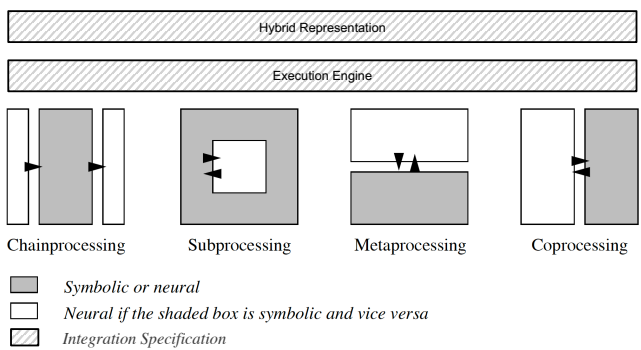

Figure 1: Integration specification and execution engine on top of Hilario's [1995] integration modes

## 2 A Hybrid Representation

To address the limitations discussed in Section 1, we propose to embrace a hybrid representation capable of combining specific features. This representation is based on the traditional fundamentals of nodes and links. There are different types of nodes and links. A collection of information units composes a terminal node. The exact notion of what constitutes an information unit is part of the node definition and depends on its specialization. In this sense, a terminal node may represent a concept, a neuron, an executable code, and media content. A context node is a composite node (or a set) that may contain links, terminal nodes, and nested composite nodes. Links define relationships among nodes' anchors. In this paper, we focus on SPO (subject-predicate-object) links capable of specifying facts such as in RDF (Resource Description Framework) and causal links where there are conditions and actions involved in the relation. Causal links allow the definition of executing the representation, while the SPO links allow the reasoning on the symbolic and sub-symbolic links. Anchors can represent the entire information of a node or its fragments in time and space.

Figure 2 shows the components (1a) that integrate the neural network topology represented as elements that connect to an Image Classifier A (1b). We begin by connecting two entities with that element: an Image (which the Image Classifier consumes) and a concept Pet (which the Image Classifier predicts) (2b). Next, we connect several entities that provide human-readable representations (labels) of their outcomes (2b). In this example, we break the network topology into smaller parts, such as a residual block [Kaiming et al., 2016] entity, a classifier layer, a skip connection entity, and a feature extractor (3b). Although the process of identifying specific structures inside a network is out of the scope of this paper, here we show the feasibility of the proposed representation to handle that type of structure. Finally, the last row of Figure 2 shows how one could link a set of rules to a neural model. In this specific step, we are using causal executable links. In this sense, we can specify a condition waiting for the feature extraction anchor to emit a signal (i.e. results in the feature extraction segment of the neural network). This result is then considered in the computation of the rules specified in the context node Rules. Note that, the causal link may specify a set of conditions and a set of actions to be triggered when the conditions are satisfied. In the example of Figure 2, the rules computation triggers another causal executable link, sending the results from this computation as an input to the classifier segment of the neural network. It is also important to highlight how the middleware is in charge of understanding the specifications, according to the definitions of the hybrid representation to deliver the appropriate content to the respective processors (e.g., neural network processors, rule and processors).

The right column of Figure 2 shows a contribution of the proposed representation: how a neural network model (as the one described in the left column) can be modeled on the proposed Hybrid representation using graph elements. Notice that, such representation differs from other graphs fundamentally; current AI frameworks and tools, e.g., Tensor-Flow, map each node to a specific function, e.g., convolution, and crop. Structuring the representation through hybrid entities aims at representing AI models in general, allowing describing not only models (neural and symbolic) but the integration between models and their corresponding processors. Moreover, the entities also support representing models' workflows, while supporting their traceability to keep track of every change applied to the models and their related entities (e.g., data, concepts, and so forth) since its creation, that is, the lifecycle of the models.

## 3 Holistic Modeling through a Hybrid Representation

In this section, we illustrate the value of the proposed representation by modeling a use case. Figure 3 illustrates this model example. First, it describes how we connect essential elements to a neural network, such as the datasets that provide training, validation, and testing sets for the network. In particular, we note how easy it is to manipulate a neural network that identifies both cats and dogs with no need for prior experience with neural network modeling: the user connects two entities named Dog and Cat that are able to detect and produce labels whenever a dog or a cat, respectively, is found on a given input image. Additionally, the representation can reside in a searchable database; a user can search for existing entities by name (such as Lion or Horse) and extend the neural network so that it includes these two animals on future image classifications.

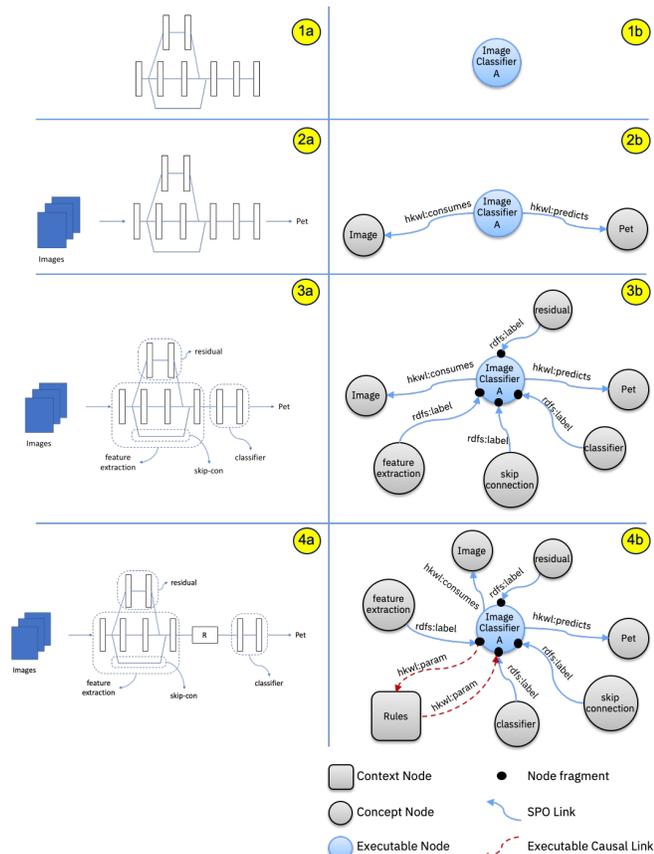

Figure 2: Modeling Neural-Symbolic Integration

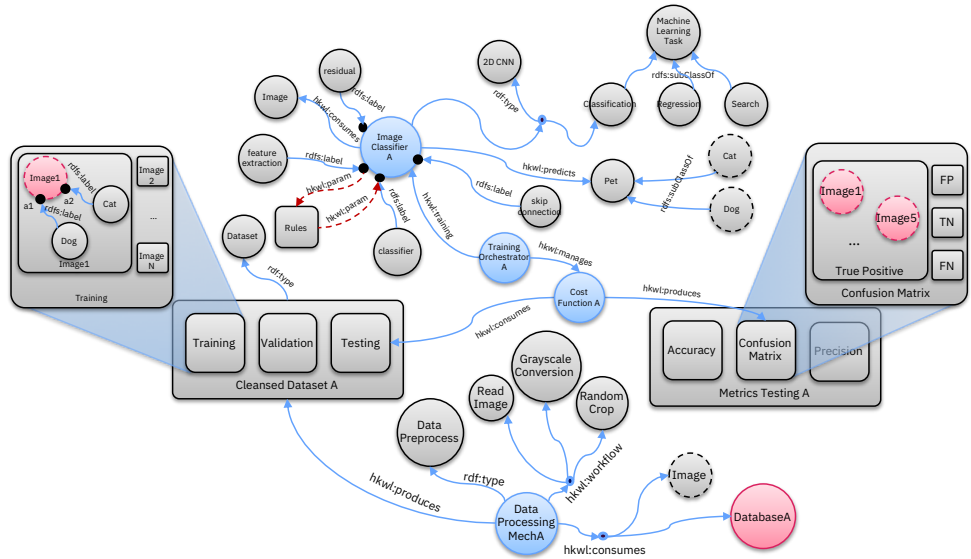

Figure 3: Use case representation

Figure 3 also depicts how to represent a machine learning experiment comprising both a dataset and the classifier specifications. For the dataset, the representation includes the dataset definition, the composition of transformations, and its association with testing metrics. The image classifier A contains a set of rules and connectionist components (also mentioned in Figure 2). Finally, the training orchestrator A links both dataset and model with cost function A; the cost function A produces metrics such as the Accuracy or Precision of the model supported by a Confusion Matrix entity, for example.

The remaining of Figure 3 shows how elements in the representation can be used (or reused, as in the Cat and Dog subclasses of the Pet element output by the Image Classifier A) to enrich the model and to describe the overall pipeline of machine learning models. This hybrid representation may be applied to support different types of queries, such as:

1. SELECT (context) node WHERE node IS dataset
2. SELECT node WHERE node IS dataset AND dataset CONTAINS cat images
3. SELECT DISTINCT dataset WHERE image FROM dataset AND image HAS cat
4. SELECT code WHERE architecture HAS code AND code HAS convolutional_layer
5. SELECT DISTINCT trained_model WHERE training_orchestrator TRAINS trained_model AND training_orchestrator CONSUMES dataset AND image HAS pet AND image FROM dataset AND training_orchestrator PRODUCES testing_metrics AND test FROM testing_metrics AND test.accuracy >= 0.9

## 4  Final Remarks

In this paper, we present our position for a neural-symbolic integration strategy. The strategy is based on the dynamicity of having an execution engine acting as a middleware, capable of interpreting high-level descriptions according to a well-defined representation to orchestrate multiple processors. The representation aims at expressing AI models in general, allowing describing not only knowledge models but the integration between these models and their corresponding processors. Our current implementation of the execution engine is already capable of parsing and executing representations such as the one illustrated in Figure 3. Queries such as the ones introduced in Section 3 are also supported. Results of our current experiments with the execution engine are encouraging us to move forward with this research initiative. Specifically, we intend to pursue the creation of mechanisms to automatic generate the models by combining fragments through meaning specification.